\documentclass[10pt]{article}
\usepackage{sober}
\usepackage{times}
\usepackage{latexsym}
\usepackage{epsfig}
\usepackage[latin1]{inputenc}

\begin{document}
\renewcommand{\thefootnote}{\fnsymbol{footnote}}
\begin{center}
{\large Acoustic Sensitive Language Model Perplexity for Automatic Speech Recognition\\[1ex]}
Ciprian Chelba\footnote{Attending Author\\
{\bf Category:} Estimation, Prediction, and Sequence Modeling\\
{\bf Preference:} 1. Poster 2. Oral}\\
Microsoft Research\\
One Microsoft Way\\
Redmond, WA 98052\\
{\tt chelba@microsoft.com}
\end{center}
\renewcommand{\thefootnote}{\arabic{footnote}}
\vspace{2.4ex}

Traditional evaluation of language models (LM) for automatic speech recognition (ASR) uses either the information theoretic -motivated \emph{perplexity (PPL)} or the \emph{word error rate (WER)} --- measured by plugging the model in a speech recognizer.

It is a well known fact that WER and PPL and poorly correlated. The main reason is probably the fact that PPL measures the predictive power of the LM on correct text, whereas at recognition time the LM needs to \emph{discriminate} between alternates suggested by the \emph{acoustic model} used in the recognizer. Since the LM is estimated using maximum-likelihood methods on correct (well-formed) sentences, it is poorly suited for discriminating among the candidates proposed by the acoustic model as likely candidates.

We propose a new evaluation metric for LMs that takes into account the coupling between language model and acoustic model in a given ASR system. The new metric, ``acoustic model -sensitive'' perplexity (AMS-PPL), aims at allowing one to optimize the LM parameters such that it performs best when used with a given acoustic model.

The underlying main idea is to estimate the conditional cross-entropy $H(W|A)$ for the correct word sequence $W$ when the acoustic signal to be decoded was $A$.

\section{Probability Model}
\label{sec:prob_model}

Let us assume that the correct word sequence $W = w_1 \ldots w_n$ is present\footnote{For the case where it isn't, one can use the lowest WER word sequence present in the lattice; ties are broken arbitrarily} in the ASR lattice~\cite{htk_book} $LAT(A)$, obtained by a given ASR system running acoustic model $P_{AM}(\cdot)$ and language model $P_{LM}(\cdot)$ on utterance $A$. We wish to calculate the conditional probability $P_{LAT}(w|w_1^k, A)$ for all vocabulary words $w\ \in\ \mathcal{V}$ and all prefixes $w_1^k$ of $W$.

Let us denote with $w_1^k \doteq w_1 \ldots w_k$ the $k$-length prefix of $W$, and with $END(w_1^k, LAT(A))$ the set of end nodes for partial paths whose word sequence is $w_1^k$:
\begin{eqnarray}
  \lefteqn{END(w_1^k, LAT(A)) \doteq \{n \in LAT(A) : (\exists)\ \pi,} \label{eq:end}\\
  & & start\_node(\pi) = s, end\_node(\pi) = n, word(\pi) = w_1^k\}\nonumber
\end{eqnarray}
where $s$ denotes the start node of $LAT(A)$ and the $word(\cdot)$ operator extracts the words on a given path. With this notation:
\begin{eqnarray}
  P_{LAT}(w|w_1^k, A) = \sum_{n \in END(w_1^k, LAT(A))} P_{LAT}(n|w_1^k, A) \cdot P_{LAT}(w|n, w_1^k, A) \label{eq:sum}
\end{eqnarray}
Let us denote the vector of acoustic observations between two nodes $n1, n2$ with $A_{n1}^{n2}$. With this notation we have $A = A_s^e$, and $A = A_s^n \parallel A_n^e$ for every node in $LAT(A)$ that is different from either the start or the end one; $\parallel$ denotes concatenation.

\subsection{Prefix-Sensitive Node Posteriors}

The first factor in the inner term of the summation in Eq.~(\ref{eq:sum}) becomes:
\begin{eqnarray*}
  P_{LAT}(n|w_1^k, A_s^n \parallel A_n^e) = {P_{LAT}(n, w_1^k, A_s^n \parallel A_n^e) \over \sum_{o \in END(w_1^k, LAT(A))} P_{LAT}(o, w_1^k, A_s^o \parallel A_o^e)}
\end{eqnarray*}
Since the underlying model is a Hidden Markov Model (HMM), the probability needed above can be rewritten as:
\begin{eqnarray*}
  P_{LAT}(n, w_1^k, A_s^n \parallel A_n^e) = \underbrace{P_{LAT}(n, w_1^k, A_s^n)}_{\alpha(n,w_1^k)} \cdot \underbrace{P_{LAT}(A_n^e|n)}_{\beta(n)}
\end{eqnarray*}
where:
\begin{eqnarray*}
  \alpha(n, w_1^k) & \doteq & \sum_{{\pi:start\_node(\pi)=s, end\_node(\pi)=n, \atop word(\pi)=w_1^k}} \prod_{i=1}^{length(\pi)} p(l_i(\pi)) \\
  \beta(n) & \doteq & \sum_{{\pi:start\_node(\pi)=n, end\_node(\pi)=e \atop }} \prod_{i=1}^{length(\pi)} p(l_i(\pi))
\end{eqnarray*}
The $\beta(n)$ is the familiar backward probability at node $n$. The $\alpha(n, w_1^k)$ is a modified version of the forward probability at node n that takes into account only partial paths whose word string matches the $w_1^k$ prefix; both can be efficiently calculated using standard dynamic programming algorithms.

To complete the analogy with the standard forward/backward definitions, we can denote:
\begin{eqnarray}
  \gamma(n|w_1^k, A_s^n \parallel A_n^e) & \doteq & P_{LAT}(n|w_1^k, A_s^n \parallel A_n^e) \label{eq:gamma}
\end{eqnarray}

\subsection{Acoustic Sensitive Word Probability}

For a given link $l$ leaving node $n$ we can write:
\begin{eqnarray}
  P_{LAT}(l|n, A_n^e) & = & {P_{LAT}(l, A_n^e|n) \over \sum_{q:start\_node(q)=n} P_{LAT}(q, A_n^e|n)} \nonumber \\
  & = & {P_{LAT}(l) \cdot \beta(end\_node(l)) \over \sum_{q:start\_node(q)=n} P_{LAT}(q) \cdot \beta(end\_node(q))} \label{eq:link_prob}
\end{eqnarray}
Making use of the fact that the underlying model is an HMM, the second factor in the inner term of the summation in Eq.~(\ref{eq:sum}) can be re-written as:
\begin{eqnarray}
  P_{LAT}(w|n, w_1^k, A) & = & P_{LAT}(w|n, A_n^e) \nonumber \\
  & = & \sum_{l:word(l)=w} P_{LAT}(l|n, A_n^e) \nonumber \\
  & = & \sum_{l:start\_node(l)=n} P_{LAT}(l|n, A_n^e) \cdot \delta(w, word(l)) \label{eq:word_prob}
\end{eqnarray}

To summarize:
\begin{eqnarray*}
  P_{LAT}(w|w_1^k, A) & = & \sum_{n \in END(w_1^k, LAT(A))} \gamma(n, w_1^k) \cdot P_{LAT}(w|n, w_1^k, A) \\
  P_{LAT}(w|n, w_1^k, A) & = & {\sum_{l:start\_node(l)=n} P_{LAT}(l) \cdot \beta(end\_node(l)) \cdot \delta(w, word(l)) \over \sum_{l:start\_node(l)=n} P_{LAT}(l) \cdot \beta(end\_node(l))}
\end{eqnarray*}
where $\gamma(n, w_1^k)$ is calculated according to Eq.~(\ref{eq:gamma}), and the probability of a link $P_{LAT}(l)$ is calculated using standard methods in ASR.

\section{Acoustic Model Sensitive Perplexity}

The probability $P_{LAT}(w|w_1^k, A)$ can be used to calculate the \emph{acoustic model -sensitive perplexity} (AMS-PPL). The lattices for the test data need to be generated/rescored with the LM to be evaluated, after which one can calculate the AMS-PPL:
\begin{eqnarray}
  AMS-PPL = \exp \left[ -1/n \cdot \sum_{k=0}^{n-1} \log P_{LAT}(w_{k+1}|w_1^k, A) \right] \label{eq:ams_ppl}
\end{eqnarray}

The same metric can be used for discriminative (acoustic sensitive) LM training~\cite{jelinek95}, under the assumption that there is a sufficient amount of parallel --- text and speech --- training data, thus enabling one to generate lattices for the LM training data.

\bibliographystyle{plain}
\bibliography{mainbibfile}

\end{document}